\documentclass[conference]{IEEEtran}
\IEEEoverridecommandlockouts
\usepackage{cite}
\usepackage{amsmath,amssymb,amsfonts}
\usepackage{algorithmic}
\usepackage{graphicx}
\usepackage{textcomp}
\usepackage{xcolor}
\usepackage{enumitem}
\usepackage{multirow}
\usepackage{array}

\def\BibTeX{{\rm B\kern-.05em{\sc i\kern-.025em b}\kern-.08em
    T\kern-.1667em\lower.7ex\hbox{E}\kern-.125emX}}
\begin{document}

\title{A Weighted Vision Transformer-Based Multi-Task Learning Framework for Predicting ADAS-Cog Scores\\
}

\author{\IEEEauthorblockN{Nur Amirah Abd Hamid}
\IEEEauthorblockA{\textit{School of Digital Science} \\
\textit{Universiti Brunei Darussalam}\\
Bandar Seri Begawan, Negara Brunei Darussalam \\
mirahhamid5496@gmail.com}
\and
\IEEEauthorblockN{Mohd Ibrahim Shapiai}
\IEEEauthorblockA{\textit{Malaysia-Japan International Institue of Technology} \\
\textit{Universiti Teknologi Malaysia}\\
Kuala Lumpur, Malaysia\\
md\_ibrahim83@utm.my}
\and
\IEEEauthorblockN{Daphne Teck Ching Lai}
\IEEEauthorblockA{\textit{School of Digital Science} \\
\textit{Universiti Brunei Darussalam}\\
Bandar Seri Begawan, Negara Brunei Darussalam \\
daphne.lai@ubd.edu.bn}
}

\maketitle

\begin{abstract}
Prognostic modeling is essential for forecasting future clinical scores and enabling early detection of Alzheimer’s disease (AD). While most existing methods focus on predicting the ADAS-Cog global score, they often overlook the predictive value of its 13 sub-scores, which reflect distinct cognitive domains. Some sub-scores may exert greater influence on determining global scores. Assigning higher loss weights to these clinically meaningful sub-scores can guide the model to focus on more relevant cognitive domains, enhancing both predictive accuracy and interpretability. In this study, we propose a weighted Vision Transformer (ViT)-based multi-task learning (MTL) framework to jointly predict the ADAS-Cog global score using baseline MRI scans and its 13 sub-scores at Month 24. Our framework integrates ViT as a feature extractor and systematically investigates the impact of sub-score-specific loss weighting on model performance. Results show that our proposed weighting strategies are group-dependent: strong weighting improves performance for MCI subjects with more heterogeneous MRI patterns, while moderate weighting is more effective for CN subjects with lower variability. Our findings suggest that uniform weighting underutilizes key sub-scores and limits generalization. The proposed framework offers a flexible, interpretable approach to AD prognosis using end-to-end MRI-based learning. (Github repo link will be provided after review)
\end{abstract}


\section{Introduction}
Prognostic modeling plays a critical role in forecasting future clinical scores, offering a more comprehensive view of Alzheimer's disease (AD) progression and its underlying patterns. Unlike classification tasks that merely determine diagnostic label (AD, cognitive normal (CN), mild cognitive impairment (MCI), prognostic models aim to anticipate the trajectory of cognitive decline—capturing the progression of AD. Tools such as the Alzheimer's Disease Assessment Scale – Cognitive Subscale (ADAS-Cog) are widely recognized for their strong correlation with AD severity, as they reflect behavioral and cognitive impairments associated with the disease \cite{RN86}. Therefore, accurate prediction of these clinical scores can enhance our understanding of how brain changes relate to cognitive deterioration, providing valuable insights for long-term prognosis.

Relatively few studies address the challenge of predicting continuous clinical scores, in part due to the inherent difficulty of regression tasks and the need for robust predictive models \cite{RN48}. A major gap in current AI research on AD lies in the limited investigation into how changes in brain structure, seen in MRI scans, relate to future cognitive decline. Most existing studies focus on predicting current clinical global scores, using them as benchmarks for evaluating model performance or feature extraction quality \cite{RN3}.
Conversely, far fewer studies attempt to predict future clinical global scores, despite their crucial role in early diagnosis and tracking the progression of AD over time. \cite{RN16, RN8, RN101, RN102}. Some recent approaches have adopted multi-task learning (MTL) strategies that aim to improve predictive performance by jointly learning related tasks—most commonly disease classification and global scores regression tasks \cite{RN16, RN8, RN101}. Most existing works focus solely on forecasting the global score, often overlooking the individual sub-scores that comprise it. These sub-scores, which assess various cognitive abilities like memory, language, and attention, contain rich localized information that can enhance the model’s capacity to capture relevant structural biomarkers and support accurate global score predictions. Whilst, global scores provide only a high-level summary, whereas sub-scores reflect diverse cognitive domains linked to specific brain regions. The list of sub-questions are as follows: Q1 (Word Recall), Q2 (Commands), Q3 (Constructional Praxis), Q4 (Naming Objects and Fingers), Q5 (Ideational Praxis), Q6 (Orientation), Q7 (Word Recognition), Q8 (Remembering Test Instructions), Q9 (Spoken Language Ability), Q10 (Comprehension of Spoken Language), Q11 (Word-Finding Difficulty in Spontaneous Speech), Q12 (Delayed Word Recall), Q13 (Digit Cancellation) \cite{mohs1997adascog13}. It is important to highlight that among the 13 ADAS-Cog sub-scores, several may have a stronger influence or greater predictive value in determining the global score. Therefore, assigning greater focus to these significant sub-scores by adjusting their weights during training to penalize errors more heavily—can guide the model to prioritize clinically meaningful cognitive domains, ultimately improving both predictive accuracy and model interpretability. To date, no study has explored a weighted-based MTL framework that simultaneously predict 13-items ADAS-Cog sub-score to explicitly predict future global scores.
Additionally, the current study builds upon our previous MTL framework, which jointly predicted both the ADAS-Cog global score and its 13 sub-scores. In that work, we demonstrated that only a few sub-scores—namely Q1, Q4, and Q8—dominated the global score prediction, yet suffered from higher prediction errors due to model instability and overreliance on clinical features. 
Motivated by these findings, the current study explores a refined MTL approach that focuses solely on the 13 sub-scores, applying task-specific loss weights to emphasize clinically significant sub-scores during training. By doing so, we aim to improve the predictive capacity of MRI-only inputs and better align sub-score predictions with their known contributions to global scores, especially in more heterogeneous clinical subgroups such as MCI.

This study aims to bridge this gap by proposing a weighted Vision Transformer (ViT) based  MTL framework to forecast the ADAS-Cog global score at Month 24. The model leverages baseline MRI from baseline as input features. Our key objectives are: (1) to develop an MTL framework that jointly predicts 13 individual ADAS-Cog sub-scores, leveraging the ViT as the backbone to capture fine-grained structural features; (2) to investigate the impact of different weighting loss ($\omega$) on model performance to emphasize clinically significant sub-scores on predictive performance; and (3) to validate the proposed approach against existing methods using ground-truth clinical data.

The remainder of this paper is structured as follows: Section II details the proposed methodology. Section III presents the experimental results and discussion. Section IV concludes the paper and outlines future research directions.

\section{Methodology}

\begin{figure}
    \centering
    \includegraphics[width=1.0\linewidth]{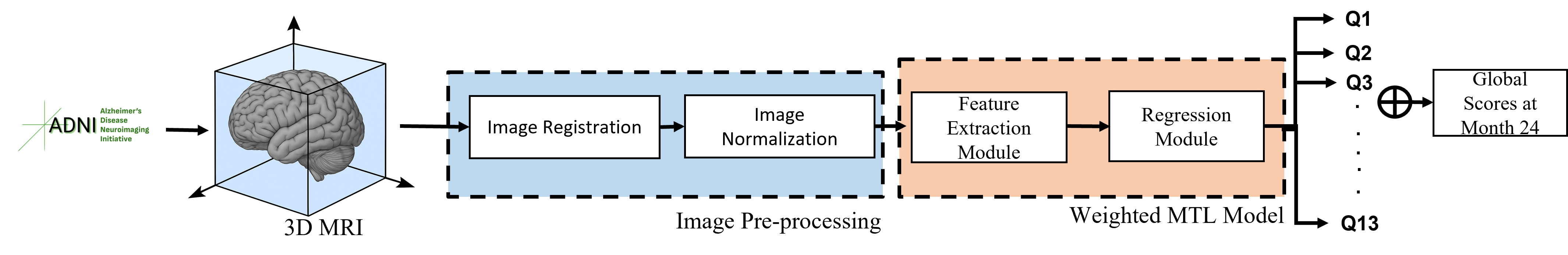}
    \caption{Proposed MTL Framework}
    \label{fig:MTL framework}
\end{figure}

The methodology in this paper involves several key steps as illustrated in Figure \ref{fig:MTL framework}: data acquisition, image preprocessing and multi-task learning model training.

\subsection{Data Acquisition}

The dataset used in this study is collected from Alzheimer’s Disease Neuroimaging Initiative (ADNI) database (https://ida.loni.usc.edu/login.jsp). ADNI dataset provides comprehensive structural MRI (sMRI) scans associated with neuropsychological tests scores from NC, MCI and AD patients, ensuring the representative sample for training and evaluation.

\begin{table}[h]
    \centering
    \caption{Demographic Characteristics}
    \label{tab:demographics}
    {\large
    \resizebox{1.0\columnwidth}{!}{
    \begin{tabular}{|>{\centering\arraybackslash}p{1cm}|c|c|c|c|}
        \hline
        \multirow{2}{*}{Group} & \multirow{2}{*}{Subject (n)} & \multicolumn{3}{c|}{Demographic Characteristics} \\ \cline{3-5}
        & & Gender (Male/Female) & Age (mean ±SD) & ADAS-Cog (mean±SD) \\ \hline
        NC & 163 & 83/80 & 76.0 ± 5.2 & 8.9±4.2 \\ \hline
        MCI & 95 & 64/31 & 75.0 ± 7.3 & 13.6±4.3 \\ \hline
    \end{tabular}
    }}

\end{table}

The study involves longitudinal follow-up data spanning up to two years, with observations at baseline, and 24 months. Relevant subjects and data are selected based on demographic characteristics, neuropsychological cognitive test, particularly ADAS-Cog, and MRI acquisition properties.  On MRI acquisition properties, we selected the data with MRI imaging protocol: T1-weighted structural sequences, a volumetric 3D MPRAGE, sagittal acquisition plane and using 1.5T Siemens System. Demographic characteristics such as age, gender, ADAS-Cog scores and research group reported in Table \ref{tab:demographics}. This study considered global ADAS-Cog scores range from 0 to 20, with a cut-off 20 indicating moderate AD, as suggested by \cite{RN71}, and total of 13 sub-scores. Our study focuses on group NC and MCI with the global score with range 20.

\subsection{Image Preprocessing}

In this study, we focused exclusively on data from the ADNI 1 cohort. We included a total of 258 MRI scans from baseline only, with ADAS-Cog global scores at Month 24 and 13 individuals sub-scores at Month 24. As the period within 24 months is important for choosing participants and planning clinical trials, so it’s very valuable to have accurate predictions of future clinical scores during this time \cite{Utsumi2018PersonalizedGP}.
The anatomical and intensity heterogeneity in MRI data can interfere the consistency accuracy. Hence, we performed two prominent preprocessing steps to reduce variability and improving the reliability of downstream MRI analysis:

\begin{enumerate}[label=\alph*.]

    \item Image registration ensures that the MRI scans are aligned to a common anatomical space,

    \item Image normalization adjusts the image intensities to ensure consistency across scans, which helps reduce inter-subject variability.
\end{enumerate}
Advanced Normalization Tools (ANTS) \cite{RN94} was used to perform image registration and normalization.

\subsection{Multi-Task Learning Model (MTL)}

Our proposed multi-task learning (MTL) model consists of two primary components: (1) a Feature Extraction Module and (2) a Regression Module. The model is designed to perform multiple related regression tasks simultaneously—specifically, predicting the 13 individual ADAS-Cog sub-scores at Month 24. By doing so, it implicitly enables the prediction of the global ADAS-Cog score, which is a composite of these sub-scores. The MTL framework is adopted based on the premise that shared learning across related tasks can enhance the extraction of common, informative features from MRI scans. This approach improves the model’s predictive accuracy and robustness across tasks by leveraging inter-task dependencies.

\subsubsection{Feature Extraction Module}

The Feature Extraction Module is a crucial step that captures salient and representative features from MRI data. In this study, we utilize ViT, a transformer-based architecture—due to its ability to model global spatial relationships across the image without relying on a fixed receptive field \cite{RN75}. This characteristic is particularly beneficial in medical imaging, where understanding the broader anatomical context is vital for accurate disease characterization.

\subsubsection{Regression Module}
Our regression module is responsible for learning to predict the 13 ADAS-Cog sub-scores at Month 24. Each sub-score is treated as a separate regression task. Following the feature extraction step, the shared representation is passed through 13 task-specific fully connected (FC) layers, enabling both joint representation learning and task-specific fine-tuning.

\subsection{Proposed weighted ViT based MTL Framework}

\begin{table}[h]
    \centering
    \caption{Correlation between sub-scores and global scores}
    \label{tab:subscores correlation}
    {\small
    \resizebox{1.0\columnwidth}{!}{
    \begin{tabular}{|>{\centering\arraybackslash}p{4cm}|>{\centering\arraybackslash}p{4cm}|}
    \hline
    \textbf{Sub-scores} & \textbf{r ±SD }  \\ \hline
    \textbf{Q1} & \textbf{0.77 ±1.34} \\ \hline
    Q2 & 0.13 ±0.28 \\ \hline
    Q3 & 0.16 ±0.52 \\ \hline
    \textbf{Q4} & \textbf{0.84 ±2.11} \\ \hline
    Q5 & 0.20 ±0.26 \\ \hline
    Q6 & 0.04 ±0.23  \\ \hline
    Q7 & 0.34 ±0.46  \\ \hline
    \textbf{Q8} & \textbf{0.70 ±2.24} \\ \hline
    Q9 & 0.14 ±0.09 \\ \hline
    Q10 & 0.12 ±0.09  \\ \hline
    Q11 & 0.19 ±0.32  \\ \hline
    Q12 & 0.02 ±0.12 \\ \hline
    Q13 & 0.35 ±0.80 \\ \hline
     
    \end{tabular}
    }}
\end{table}

To further enhance the model's focus on clinically important cognitive domains, we propose a  weighted ViT based MTL framework. In this approach, we apply a weighted loss function that assigns higher importance to certain sub-scores. Specifically, we first compute the Pearson correlation between each sub-score and the global ADAS-Cog score at Month 24. Based on the observed high correlations, sub-scores Q1, Q4, and Q8 are identified as key components to the global score (see Table \ref{tab:subscores correlation}) and are thus assigned higher weights during training.

The total loss function used in the our setup combines the weighted loss for 13 ADAS-Cog sub-scores ($Q_1$ to $Q_{13}$) and defined in Equation (1) and Equation (2). This loss function encourages the model to prioritize accurate prediction of $Q_1$, $Q_4$, and $Q_8$ more than the other sub-scores, while still learning to predict the global score. This reflects domain knowledge where certain cognitive tests may carry higher clinical significance.

\begin{equation}
\mathcal{L}_{\text{total}}(X, Y) = \sum_{j=1}^{13} \mathcal{L}_{\text{mse (sub-scores)}}^{j}
\end{equation}

where ${L}_{\text{mse (sub-scores)}}^{j}$ is the MSE loss for the ${j}^{th}$ sub-score.

\begin{equation}
\mathcal{L}_{\text{mse (sub-scores)}}^{j} = \frac{1}{B} \sum_{i=1}^{B} w_j \left( y_{ij} -\hat{y}_{ij} \right)^2
\end{equation}

where,

\begin{itemize}
  \item $B$: The batch size (number of subjects in a mini-batch).
  \item $y_{ij}$: The \text{true} value of the $j^\text{th}$ sub-score for subject $i$.
  \item $\hat{y}_{ij}$: The \text{predicted} value of the $j^\text{th}$ sub-score for subject $i$.
  \item $w_j$: A \text{task-specific weight} for sub-score $j$, used to emphasize clinically significant sub-scores.
\end{itemize}

\begin{table*}[t]
\centering
\caption{Comparison of weighting strategies applied to ADAS-Cog sub-scores in multi-task learning.}
\label{tab:weighting_strategies}
\begin{tabular}{|l|c|c|c|c|p{4cm}|}
\hline
\textbf{Configuration} & \textbf{Q1/Q4/Q8 ($\omega$)} & \textbf{Other Sub-scores ($\omega$)} & \textbf{Q1+Q4+Q8 ($\omega$ \%)} & \textbf{Others ($\omega$ \%)} & \textbf{Description} \\ \hline
Uniform weighted ViT & 1 & 1 & 23.1\% & 76.9\% & Equal weight to all 13 sub-scores (no emphasis). \\ \hline
Moderate Weighted ViT & 0.160 & 0.052 & 48.0\% & 52.0\% & Q1, Q4, Q8 given moderate importance (16.0\%), others at 5.2\%. \\ \hline
Strong Weighted ViT & 0.320 & 0.004 & 96\% & 4\% & Q1, Q4, Q8 highly emphasized (32.0\%), others minimally weighted. \\ \hline
\end{tabular}
\end{table*}

\noindent In our ablation study, we investigated and compared three settings as shown in Table \ref{tab:weighting_strategies}. Note that, we randomly set-up the $\omega$ value for the three settings.

\subsection{Model Training and Evaluation}

We trained our weighted ViT based multi-task learning (MTL) model using preprocessed and normalized baseline MRI scans to predict 13 individuals sub-scores at Month 24. Additionally, we investigated the impact of using different values of weighting loss on a specific task on the model's performance in predicting future clinical scores.
The training process involved the following steps:
\begin{enumerate}
    \item Data Splitting: The dataset was divided into training and validation sets with ratio 80:20 at the subject level to prevent data leakage and ensure robust model evaluation.
    \item Model Training: The model was trained on the training set using an Adam optimizer and a combined loss function tailored for multi-output regression. We set up the experiments based on Table \ref{tab:weighting_strategies}.
    \item Evaluation Metrics: Model performance was assessed using mean absolute error (MAE), root mean squared error (RMSE), and Pearson correlation coefficient (r). The Pearson correlation 
    as used to quantify the linear relationship between the actual and predicted scores, with higher values indicating stronger predictive performance.
\end{enumerate}

\section{Results and Discussion}

\subsection{Global Score at Month 24 Prediction}

\begin{table}[h]
    \centering
    \caption{Weighted ViT based MTL Framework: Performance evaluation based on weighting strategies.}
    \label{tab:ablation study}
    {\large
    \resizebox{1.0\columnwidth}{!}{
    \begin{tabular}{|>{\centering\arraybackslash}p{3cm}|>{\centering\arraybackslash}p{4cm}|>{\centering\arraybackslash}p{2cm}|r|r|>{\centering\arraybackslash}p{1cm}|}
    \hline
    \textbf{Architecture} & \textbf{$\omega$ } & \textbf{Input data} & \textbf{MAE} & \textbf{RMSE} & \textbf{r} \\ \hline
    Strong weighted ViT (ours) & Q1, Q4, Q8 = 0.32 (32\%) Otherwise = 0.004 (0.4\%) & Baseline MRI & 4.49 & 5.29 & 0.21 \\ \hline
    Moderate weighted ViT (ours) & Q1, Q4, Q8 = 0.16 (16\%) Otherwise = 0.004 (0.5\%) & Baseline MRI & 4.52  & 5.16 & 0.24 \\ \hline
    Uniform weighted ViT & Equivalent & Baseline MRI & 4.58 & 5.28 & 0.13 \\ \hline
    \end{tabular}
    }}
\end{table}

\begin{table}[h]
    \centering
    \caption{Weighted ViT based MTL Framework: Comparison of Performance evaluation based on group}
    \label{tab:comparison table}
    {\large
    \resizebox{1.0\columnwidth}{!}{
    \begin{tabular}{|>{\centering\arraybackslash}p{2cm}|>{\centering\arraybackslash}p{4cm}|>{\centering\arraybackslash}p{2.5cm}|r|r|>{\centering\arraybackslash}p{1cm}|}
    \hline
    \textbf{Group} & \textbf{Architecture} & \textbf{Input data} & \textbf{MAE} & \textbf{RMSE} & \textbf{r} \\ \hline
    \multirow{2}{*}{CN} 
    & Strong weighted ViT (ours) & Baseline MRI & 3.94 & 4.74 & 0.10 \\ \cline{2-6}
    & Moderate weighted ViT (ours) & Baseline MRI & 4.08 & 4.62 & 0.30 \\ \cline{2-6}
    & Dirty Model \cite{RN101} & Baseline MRI & 3.18 & N/A & 0.08 \\ \hline
    \multirow{2}{*}{MCI} 
    
    & Strong weighted ViT (ours) & Baseline MRI & 5.32 & 6.02 & 0.27 \\ \cline{2-6}
    & Moderate weighted ViT (ours) & Baseline MRI & 5.18 & 5.87 & 0.15 \\ \cline{2-6}
    & Dirty Model \cite{RN101} & Baseline MRI & 5.06 & N/A & 0.37 \\ \hline
    \end{tabular}
    }}
\end{table}

Table \ref{tab:ablation study} presents a focused ablation study evaluating the impact of different weighting strategies on our ViT-based MTL model using baseline MRI as the sole input modality. All configurations share the same regression module composed of linear layers, and differ only in the weighting scheme applied to the sub-score loss components.

Results show that both weighted configurations outperformed the uniform weighted ViT baseline. The strong-weight configuration (Q1=Q4=Q8=0.32) yielded MAE: 4.49, RMSE: 5.29, and Pearson r: 0.21, while the moderate-weight setup (Q1=Q4=Q8=0.16) achieved MAE: 4.52, RMSE: 5.16, and r: 0.24. In contrast, the uniform weighted ViT model resulted in MAE: 4.58, RMSE: 5.28, and a significantly lower correlation (r = 0.13).
These findings underscore the effectiveness of selectively emphasizing clinically significant sub-scores during training. By prioritizing specific sub-scores—particularly Q1 (Word Recall), Q4 (Delayed Word Recall), and Q8 (Word Recognition)—the model not only improves regression accuracy but also enhances the correlation between predicted and actual global ADAS-Cog scores. This improvement is especially meaningful in MRI-only scenarios, where the 3D MRI scan is complex and learning discriminative features is inherently more difficult.
The application of targeted weighting demonstrates that Q1, Q4, and Q8 play an important role in predicting future cognitive decline from structural MRI. These sub-scores are strongly associated with memory function, suggesting that subtle changes in brain structure—reflected in MRI—are most prominently captured through memory-related subcomponents. This supports the hypothesis that MRI-based features are more predictive for cognitive domains like memory, and that strategic weighting can help the model exploit this relationship more effectively.

To evaluate the impact of subject group on prediction performance, we conducted a subgroup analysis focusing on two clinical cohorts—CN and MCI—as shown in Table \ref{tab:comparison table}. We compared the performance of our weighted ViT-based MTL models (i.e., moderate and strong configurations) with the Dirty Model proposed by Imani et al. \cite{RN101}, using baseline MRI as the sole input modality.

For the CN group, our weighted ViT models outperformed the Dirty Model in terms of Pearson correlation, with the moderately weighted model (r = 0.30) achieving superior correlation compared to both the strongly weighted model (r = 0.10) and the Dirty Model (r = 0.08). This suggests that a moderate emphasis on sub-scores Q1, Q4, and Q8 is more suitable for CN subjects, where MRI features tend to be more homogeneous and exhibit less inter-subject variability. 

In contrast, for the MCI group, although both of our models underperformed the Dirty Model (r = 0.37), the strongly weighted configuration (r = 0.27) outperformed the moderate one (r = 0.15), indicating that increased focus on memory-related sub-scores is more effective in capturing structural differences among MCI subjects.

These results indicate that targeted sub-score weighting strategies may need to be group-specific task. In the MCI group, where structural brain changes are more distinct and variable, assigning greater importance to memory-related sub-scores (Q1, Q4, Q8) leads to better correlation between predicted and true global scores. In contrast, the CN group benefits from a more balanced weighting approach, as their MRI patterns are more uniform and less pathologically distinct. In such cases, a wider range of sub-scores may contribute to cognitive variation, and overemphasizing a limited subset (e.g., Q1, Q4, Q8) may reduce correlation despite relatively low absolute prediction errors (e.g., MAE). This further illustrates that low error does not necessarily imply strong predictive alignment, especially in populations with limited structural variability.

Additionally, the performance gap between our models and the Dirty Model \cite{RN101} in the MCI group may stem from differences in how MRI features are represented. While both approaches use baseline MRI, our models utilized 3D volumetric scans on an end-to-end weighted-ViT based MTL, whereas Imani et al. \cite{RN101} extract 122 handcrafted features based on regional grey matter density. These regional features may better capture disease-relevant anatomical signals, particularly in smaller or early-stage cohorts where subtle MRI variations may be challenging to learn directly from whole 3D scans. In contrast, our end-to-end approach learns spatial features directly from the data, which, while flexible and scalable, may require larger datasets or task-specific regularization to achieve comparable interpretability and correlation.

These insights support the design of our proposed MTL framework, where the global ADAS-Cog score is not predicted directly but emerges through a learned aggregation of cognitively meaningful sub-scores. They also underscore the importance of evaluating individual sub-score predictions to interpret global scores effectively. Notably, not all sub-scores contribute equally to the final prediction—some, like Q1, Q4, and Q8, have a much greater influence than the others.This opens opportunities for future work to explore adaptive or dynamic loss weighting mechanisms that emphasize sub-scores which are both MRI-relevant and clinically informative, leading to more balanced, interpretable, and group-sensitive learning.


\section{Conclusion}

This study presented a weighted ViT based MTL framework for predicting future ADAS-Cog 13 global scores at Month 24 by leveraging both individual sub-scores and baseline MRI scans. Unlike most previous studies that focus solely on predicting the global ADAS-Cog score, our approach introduces a weighted loss formulation that selectively emphasizes clinically relevant sub-scores—specifically Q1, Q4, and Q8—based on their strong correlation with overall cognitive decline.

Through extensive ablation and subgroup analyses, we demonstrated that the effectiveness of sub-score weighting is group-dependent: strong weighting is more beneficial for the MCI group, where structural brain changes are more heterogeneous, whereas moderate weighting performs better for the CN group, where MRI variability is lower. Our findings highlight that accurate global score prediction is not solely dependent on absolute error metrics such as MAE or RMSE but also requires careful consideration of model alignment with inter-subject variability, as reflected in correlation performance. Moreover, our results suggest that conventional uniform loss weighting may underutilize informative sub-scores and lead to suboptimal model generalization, particularly in heterogeneous clinical populations.

While our end-to-end ViT approach using 3D MRI volumes offers a flexible learning framework, we acknowledge the advantages of handcrafted regional features in small-cohort or early-stage prediction settings. Future work will explore adaptive loss weighting schemes, integration of multimodal data (e.g., cognitive scores, demographics), and explainability tools to improve both model interpretability and clinical utility. Overall, this work higlights the value of incorporating clinical domain knowledge into model design and demonstrates the potential of weighted MTL strategies to enhance the sensitivity and interpretability of cognitive decline prediction from structural MRI.

\bibliographystyle{ieeetr}
\bibliography{references}

\end{document}